\documentclass[letterpaper,10pt,conference]{ieeeconf}
\IEEEoverridecommandlockouts
\usepackage{cite}
\usepackage{amsmath,amssymb,amsfonts}
\usepackage{graphicx}
\usepackage{textcomp}
\usepackage[table]{xcolor}
\usepackage{newtxtext,newtxmath}
\usepackage{booktabs}
\usepackage{multirow}
\usepackage{makecell}
\usepackage{tabularx}
\usepackage{url}
\usepackage{balance}
\usepackage{placeins}
\usepackage{stfloats}
\usepackage{float}
\usepackage{xspace}
\usepackage{hyperref}
\graphicspath{{TU/}{figures/}}
\hypersetup{
colorlinks=true,
linkcolor=black
}

\definecolor{bestblue}{RGB}{228,239,250}
\newcommand{\bestvalue}[1]{\textbf{#1}}
\newcommand{\bestmethod}[1]{\textbf{#1}}
\newcolumntype{Y}{>{\centering\arraybackslash}X}


\definecolor{spatialgreen}{RGB}{65,130,85}
\definecolor{appearorange}{RGB}{185,115,40}
\definecolor{demoblue}{RGB}{70,105,150}
\definecolor{distractgold}{RGB}{170,125,35}
\definecolor{similarred}{RGB}{170,75,65}
\definecolor{confblue}{RGB}{55,105,145}
\definecolor{rejectgold}{RGB}{165,120,25}
\definecolor{fomgreen}{RGB}{45,125,75}

\title{\LARGE \bf
Towards Fine-Grained Object Manipulation: SAM3-Guided \\
Visuomotor Policy with Persistent Memory Learning and \\ Focused Visual Conditioning
}
\newif\ifanonymous
\anonymousfalse

\ifanonymous

\author{\rule{0pt}{2.2ex}}

\else

\author{Haolong Meng$^{1,2}$, Fangbo Qin$^{*1,2,4}$, Mengchen Bai$^{1}$, Houwu Wang$^{1}$, Xiaolin Liu$^{1}$, Cirong Liu$^{3,4}$, Shan Yu$^{1,2,4}$
\thanks{This work was supported by the Brain Science and Brain-like Intelligence Technology - National Science and Technology Major Project (2025ZD0219300) and the Beijing Nova Program (202604841208). {Corresponding author: F. Qin; Email: fangbo.qin@ia.ac.cn}}%
\thanks{$^{1}$Institute of Automation, Chinese Academy of Sciences, Beijing 100190, China. 
$^{2}$School of Artificial Intelligence, University of Chinese Academy of Sciences, Beijing 100049, China. 
$^{3}$Center for Excellence in Brain Science and Intelligence Technology, Chinese Academy of Sciences, Shanghai 200031, China.
$^{4}$State Key Laboratory of Brain Cognition and Brain-Inspired Intelligence Technology, Shanghai 200031, China}%
}
\fi
\centerfigcaptionsfalse

\begin{document}
\bstctlcite{BSTcontrol}
\maketitle
\setlength{\topmargin}{-0.2in}

\begin{abstract}
\setlength{\baselineskip}{9.583333pt}
Fine-grained object (FO) manipulation requires robots to distinguish a specified FO from visually similar objects and execute actions reliably despite scene distractors. However, scene-level visual conditioning lacks explicit object selection, while category-level guidance cannot reliably distinguish FOs within the same category. We present a SAM3-guided visuomotor policy framework that addresses these challenges through persistent object memory and focused visual conditioning. First, we introduce FO Memory-driven SAM3 (FOM-SAM3), which learns reusable FO memory tokens from limited multi-view registration images, while keeping SAM3 fully frozen. Through one-vs-rest learning, these tokens encode persistent memories for localizing target FOs and rejecting similar alternatives, which can be stored in a memory bank for reuse. Second, we propose Focused Spatial-Appearance Encoding (FSAE), which combines in-FO appearance features with explicit bounding boxes to condition action policies, including Diffusion Policy (DP) and Action Chunking with Transformers (ACT). The effectiveness of the proposed FOM-SAM3 is validated on the FO-30 dataset comprising 30 physical objects across four coarse categories. Across three real-robot FO manipulation tasks, our FOM-SAM3-guided policies demonstrated robustness against distractors, discriminative ability of similar FOs, and extendibility to new FOs. \ifanonymous\else Code and videos are available at \href{https://hlmeng-casia.github.io/FOM-SAM3-Policy/}{\textcolor{blue}{\textbf{Project Page}}}.\fi
\end{abstract}

\begin{keywords}
Perception for Grasping and Manipulation, Learning from Demonstration, Visual Learning, Visuomotor Policy.
\end{keywords}

\section{Introduction}
\label{sec:introduction}

Visuomotor policies, which map visual and proprioceptive sensing to flexible actions, provide a key foundation for robotic manipulation \cite{zheng2025survey}. Imitation learning enables transferring human demonstrations to robot skills efficiently. The pioneering works studied action modeling and visual representation \cite{kim2021transformer,cao2023learning,gao2023k}, paving the way to skillful object manipulation. Action Chunking with Transformers
(ACT)~\cite{act} and Diffusion Policy (DP)~\cite{diffusion_policy} are two representative visuomotor policies and have achieved widespread adoption in recent years. The former combines action chunking with temporal ensembling to mitigate compounding errors and enable learning smooth manipulation from limited demonstrations. The latter models continuous action sequences through conditional denoising, capturing multimodal action distributions.
The following works further extended these policies through 3D vision~\cite{dp3, pointcloudmatters, diffuser_actor_3d}, model pruning~\cite{wu2025device}, multi-arm interdependencies ~\cite{pmlr-v270-lee25a}, and data scaling \cite{aloha_unleashed} among others, to improve precision, efficiency, and generalization ability.

\begin{figure}[t]
\centering
\includegraphics[width=\columnwidth]{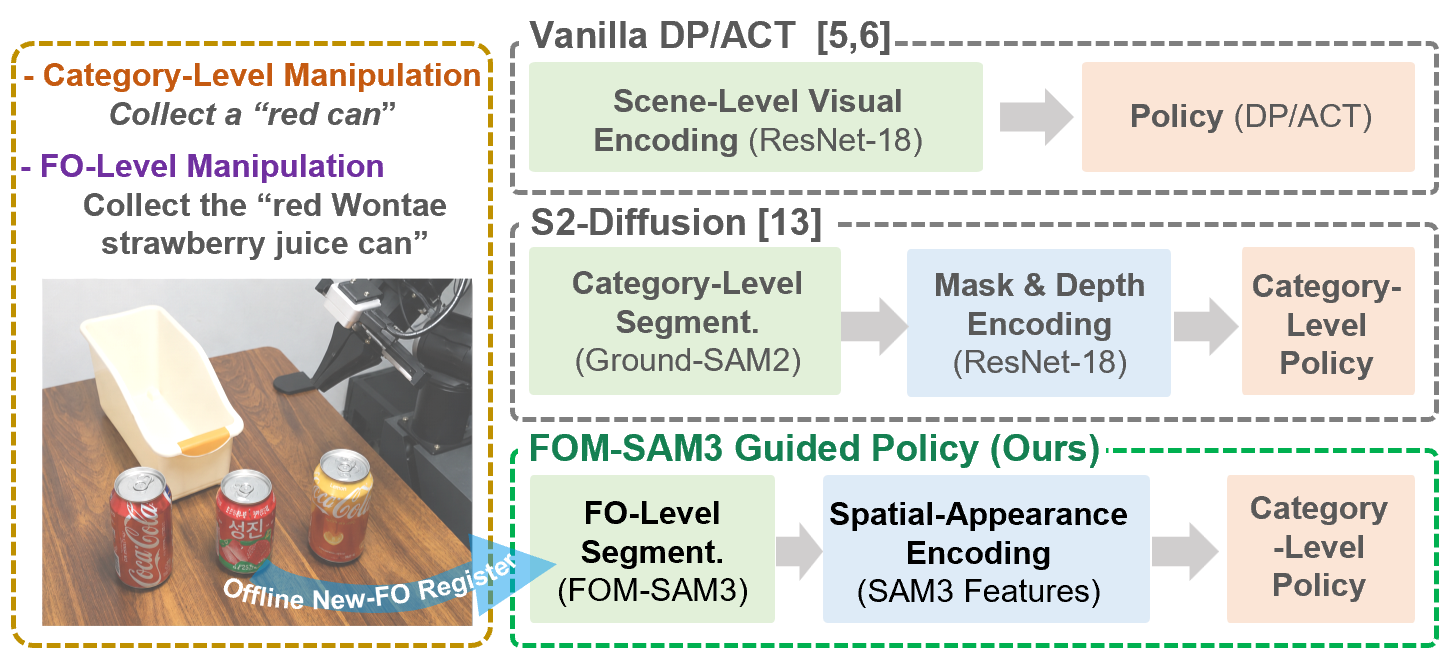}
\vspace{-20pt}
\caption{
Fine-grained object (FO) manipulation with FOM-SAM3 guided policy. In contrast to category-level object manipulation, FO manipulation requires identifying and focusing on an object with specified attributes, category, brand, and model. The current methods using scene- and category-level visual conditioning are unable to handle FO manipulation tasks. Our framework leverages SAM3 for FO-level visual guidance, which can be extended to different FOs belonging to the same coarse category via persistent memory learning.}
\label{fig:overview}
\end{figure}

In this work, we address an underexplored challenge in visuomotor policy learning: \textit{Fine-Grained Object (FO) manipulation}, which has significant value across various real-world applications. Here, the definition of FO is a target object defined by a set of stable, observable, fine-grained attributes, such as brand, model, color, pattern, or local shape, which distinguish it from other objects within the same coarse-grained category. As shown in Fig. \ref{fig:overview}, when multiple similar objects occur in the same scene, the robot needs to pick out the one defined with a fine-grained description. Common visuomotor policies do not consider the multi-target issue, and the robot relies solely on scene-level observations, without any object selection mechanism. S2-Diffusion \cite{s2diffusion} introduced Grounded-SAM2 \cite{ren2024grounded} to provide category-level object masks as visual guidance for over-instance generalization, which can filter category-irrelevant distracting objects. However, \emph{how to manipulate the target FO while minimizing distraction from scene 
variation and confusion with other FOs in the same coarse-grained category
} is still a challenging open problem for on-device visuomotor policies.

\subsection{Related Works}

\textit{1) Visual Encoders for DP and ACT:}
Visual encoders play a crucial role in visuomotor policies by extracting task-relevant information from environmental observations~\cite{sparta,spill}. The original DP and ACT employ ResNet-18 backbones for image encoding, while subsequent studies explore richer visual representations with improved geometric precision and generalization, including LoRA-adapted DINOv2 features~\cite{cage}, cross-domain visual alignment~\cite{egomimic}, 3D point-cloud encoders~\cite{pointcloudmatters,dp3,rise,idp3}, RGB-D-based lifting of visual features into 3D scene representations~\cite{diffuser_actor_3d}, and task-conditioned scene representation modulation~\cite{hypertasr}. Nevertheless, these methods condition policies on scene-level representations without explicitly decomposing observations into task-relevant objects. This may leave object features entangled with irrelevant scene content, potentially limiting robustness to distractors and generalization across layouts.

\textit{2) SAM-Guided Visuomotor Policies:}
The SAM family of visual foundation models~\cite{sam,sam2,sam3} supports visuomotor learning through transferable visual features and promptable segmentation. At the feature level, SAM provides visual priors that can be adapted to robotic observations. SAM-E~\cite{same} transfers the SAM encoder's visual priors to multi-view manipulation through parameter-efficient fine-tuning. SAM2Act~\cite{sam2act} combines SAM2 features with multi-resolution upsampling to support precise action localization. At the segmentation level, SAM facilitates the masks of relevant objects. GROOT~\cite{groot} combines SAM proposals with DINOv2-based correspondence to allow the point cloud guided policy to be generalized to new objects. HODOR~\cite{hodor} combines Grounded-SAM segmentation and DINOv2 features to construct hierarchical representations of scenes, objects, and parts. Lan-O3DP~\cite{lan_o3dp} projects SAM masks onto point clouds, focusing policy observations on task-relevant objects and reducing interference from background clutter. $S^2$-Diffusion~\cite{s2diffusion} combines Grounded-SAM2 masks with monocular depth to preserve task-relevant semantic masks and spatial depth while suppressing appearance variations, facilitating skill transfer across instances within the same category. Similarly, Ding et al.~\cite{choose_what_to_observe} use SAM3 to segment target objects and the robot gripper, rendering these entities with fixed semantic colors and incorporating monocular depth to construct task-aware observations that are robust to appearance shifts.

Although SAM can be utilized to enhance the visual conditioning, it still lacks fine-grained prompting ability. SAM3's concept prompt is limited to noun phrases and common modifiers. Lightweight adapters and fine-tuning can improve SAM's performance on specific segmentation tasks~\cite{sam2_adapter,sam3_adapter}. Prompt tuning optimizes learnable prompts while keeping pretrained model weights frozen~\cite{prompt_tuning,coop}. PTSAM~\cite{ptsam} learns prompt tokens to specialize SAM for specific segmentation tasks, with additional encoder prompt tuning mitigating domain shifts. M2C~\cite{m2c} optimizes concept embeddings within frozen SAM3 using a few annotated images. However, these methods primarily consider semantic- or concept-level adaptation, which have not explicitly explored mechanisms for enhancing discrimination among similar FOs and enabling sustained reuse of the adapted representations.

\subsection{Contributions}
Towards FO manipulation, this work addresses two key issues: 1) extending SAM3’s concept-level segmentation to FO-specific segmentation; 2) directly bridging SAM3 and an action policy without introducing an extra vision backbone. Our contributions are as below.

First, we introduce a novel FOM-SAM3-guided visuomotor policy framework for manipulating FOs in a specific operational domain. By registering new FOs in a persistent memory bank, the framework enables a shared action policy to be reused across different FOs within the same coarse category.

Second, we propose a persistent FO memory learning mechanism that keeps SAM3 fully frozen. Through one-vs-rest learning on randomly composed FO images, this mechanism adapts concept-level prompt tokens into FO-specific memory tokens for persistent storage and retrieval.

Third, we design Focused Spatial-Appearance Encoding (FSAE) to connect FOM-SAM3 with the action policy. FSAE combines explicit bounding box with implicit appearance features extracted within the target FO mask, providing visual states for FO-centric action guidance.

\section{Method}
\label{sec:method}

\subsection{Problem Definition}
\label{sec:problem_definition}

\begin{figure*}[t]
\centering
\includegraphics[width=1.0\textwidth]{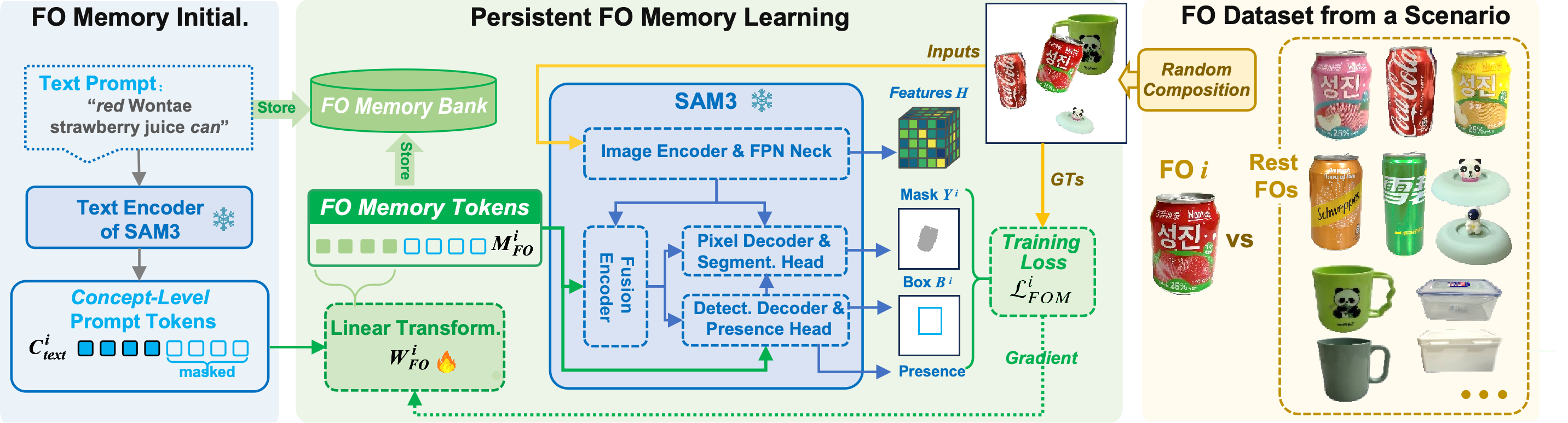}
\vspace{-20pt}
\caption{Persistent FO memory learning for SAM3.}
\label{fig:fom-learning}
\end{figure*}

We picture an application paradigm: a robot is deployed in a dedicated scenario, such as a room where the robot works every day. The robot gradually becomes familiar with the FOs
it encounters, such as
``red Wontae strawberry juice can''. Each familiar FO can be registered as a persistent memory and
indexed by an FO ID. During manipulation, an FO memory is
retrieved and fed to SAM3 ~\cite{sam3} as prompt tokens, allowing the robot to recognize
the intended FO across different scenes without external explicit per-image prompts.

The first problem is how to persistently memorize and localize an FO
with SAM3.
Given an RGB frame $I$ and the FO memory tokens $M^i_{FO}$ of the $i$-th FO,
FOM-SAM3, denoted as $\mathcal{S}$, provides the mask $Y^i$, bounding box
$B^i$, and SAM3's intermediate feature map $H$, as expressed by
\begin{equation}
\left\langle
Y^i,
B^i,
H
\right\rangle
=
\mathcal{S}
\left(
I;
M_{FO}^i
\right).
\label{eq:fomsam_problem}
\end{equation}
When multiple FOs are relevent to an action, FOM-SAM3 can run in batch mode and infer
$\{Y^i\}$ and $\{B^i\}$ from a batch of FO memory tokens
$\{M_{FO}^i\}\left(i=i_1,i_2,\ldots,i_N\right)$.
In most cases, an action refers to $N=2$ FOs: one is the active object that
the end-effector contacts, and the other is the passive object that interacts with
the active one.

The second problem is the FO-centric visual conditioning mechanism bridging
FOM-SAM3 and the action policy $\pi$.
Since the preceding FOM-SAM3 has provided FO masks and rich visual
features, they can be leveraged as the informative inputs of the encoder
$\mathcal{E}$ to obtain the
spatial-appearance representation of each FO in a camera view,
\begin{equation}
v^{i}
=
\mathcal{E}
\left(
H,
Y^{i},
B^{i}
\right).
\label{eq:fsae_problem}
\end{equation}

To enable 3D manipulation, this work adopts a compact dual-view vision setup.
The cameras $c_1$ and $c_2$ provide a static third-person view and an active
first-person view, respectively.
Thus, when two FOs are involved, the action policy is conditioned on four
FO representations and a robot state sequence $r_t$, as given by
\begin{equation}
A_t
=
\pi
\big(
v^{i_1,c_1}_t,
v^{i_2,c_1}_t,
v^{i_1,c_2}_t,
v^{i_2,c_2}_t,
r_t,
z_t
\big),
\label{eq:policy_problem}
\end{equation}
where $A_t$ is the action chunk. To indicate motion trends, $r_t$ incorporates robot states at the current and previous time steps. Both an action step and a robot state consist of the 3D position, quaternion, and opening distance of the robot gripper. 
For DP~\cite{diffusion_policy}, $z_t \sim \mathcal{N}(0,I)$
initializes the noisy action sequence. For ACT~\cite{act},
$z_t$ is the latent style variable, drawn from a parameterized normal distribution to encode multimodal demonstration styles during training and set to zero at deployment for deterministic execution.

\subsection{Persistent FO Memory Learning for SAM3}
\label{sec:fo_memory_learning}

The original SAM3~\cite{sam3} is a powerful promptable foundation model,
which takes a simple noun phrase or image exemplar as a concept prompt.
Given a concept-level prompt, such as ``red can'' or ``cup lid'', SAM3 can segment coarse-grained objects. 
However,  SAM3 struggles to interpret fine-grained descriptions such as ``red Wontae strawberry juice can'' and ``cup lid with a panda handle'' leading to confusion among similar FOs within the same coarse-grained category.

Inspired by the human ability to memorize and recognize familiar FOs, we
propose a \textit{persistent FO memory learning} mechanism built upon the
frozen SAM3.
In essence, SAM3 inference is conditioned on internal prompt tokens, which
interact extensively with the fusion encoder and decoders of SAM3.
The prompt tokens can be temporarily generated from an external text prompt
by the pretrained text encoder, or learned from limited data as a persistent
memory of a specific FO.
Therefore, we define the \textit{persistent FO memory} $M^i_{FO}$ as a set
of internal prompt tokens that directly condition the SAM3 modules to segment
the memorized FO, while excluding similar FOs belonging to the same concept.

As shown in Fig. \ref{fig:fom-learning}, to learn the effective memory of an FO, we first obtain its concept-level
prompt tokens
$C^i_{text}\in\mathbb{R}^{c\times256}$ using a noun phrase and the frozen
text encoder of SAM3.
Then the memory tokens are formed by a linear transformation of the
concept-level prompt tokens, namely,
\begin{equation}
M^{i,k}_{FO}
=
\begin{cases}
C^{i,k}_{text}W^i_{FO}, & \text{if } m_k=0,\\
C^{i,k}_{text}, & \text{if } m_k=1,
\end{cases}
\label{eq:fo_memory_tokens}
\end{equation}
where $W^i_{FO}\in\mathbb{R}^{256\times256}$ is a learnable transformation
matrix, which is initialized as an identity matrix.
$m_k$ is the padding mask determined by the text encoder, which controls the validity of the $k$-th token. During learning, the effective gradient of the loss $\mathcal{L}_{FOM}^{i}$ with respect to $M^{i,k}_{FO}$ induced by the shared matrix $W^i_{FO}$ is given by,

\begin{equation}
\left( \frac{\partial \mathcal{L}^i_{FOM}}{\partial M^{i,k}_{FO}} \right)_{\text{eff}} = \sum_{j=1,m_j=0}^c \left( C^{i,k}_{text}{C^{i,j}_{text}}^\top \right) \frac{\partial \mathcal{L}^i_{FOM}}{\partial M^{i,j}_{FO}}.
\label{eq:gradients}
\end{equation}
Thus, under gradient descent, the update to the $k$-th token $M^{i,k}_{FO}$ is a weighted aggregation of the gradients of all tokens sharing the transformation, with weights determined by the inner products between their initial tokens. The coupled token updating potentially mitigates the instability associated with independent token optimization.

Since SAM3 already possesses strong generalizable priors on objectness,
shapes, and semantics, learning $W^i_{FO}$ mainly aims to
distinguish different FOs within the same concept.
The learning for each FO is realized in a one-vs-rest paradigm.
For the $i$-th FO, while freezing the entire SAM3 and the concept-level
tokens $C^i_{text}$, we treat only the $i$-th FO as the positive target. Considering that the target FO might be present or absent in a training image, the loss from a training sample is constructed as
\begin{equation}
\mathcal{L}_{FOM}^{i}
=
\begin{cases}
\mathcal{L}_{all}^{i}, & \text{if FO-\textit{i} is present},\\
\lambda\mathcal{L}_{presence}^{i}, & \text{if FO-\textit{i} is absent}.
\end{cases}
\label{eq:fo_memory_loss}
\end{equation}
Namely, when the $i$-th FO is
present in a training image, $\mathcal{L}_{present}^{i}$ containing all the original loss items of
SAM3 training, including presence, cross-entropy, bounding box, generalized IOU, mask, and dice losses,  are used to guarantee the detection and segmentation accuracy.
When the $i$-th FO is absent, only the original presence token loss $\mathcal{L}_{presence}^{i}$ is calculated to reject the other FOs. $\lambda$ is set to 0.1 to balance the target localization and non-target rejection abilities. After learning $W^{i}_{FO}$, the token sequence $M^{i}_{FO}$ is stored
as persistent FO memory in an FO memory bank for future reuse.

The dataset for FO memory learning contains multiple FOs that appear in a
specific operational domain, which can be collected in a cumulative manner.
Each time a new FO is involved in the current scenario, we simply capture
tens of RGB frames of this FO in a pure background from various perspectives.
The FO images can be automatically annotated by foreground segmentation.
With the single-FO images and their masks, we can randomly synthesize multi-FO scenes via compositing the FO foregrounds with varying numbers, selections, scales, and poses.

\begin{figure*}[!t]
\centering
\includegraphics[width=1.0\textwidth]{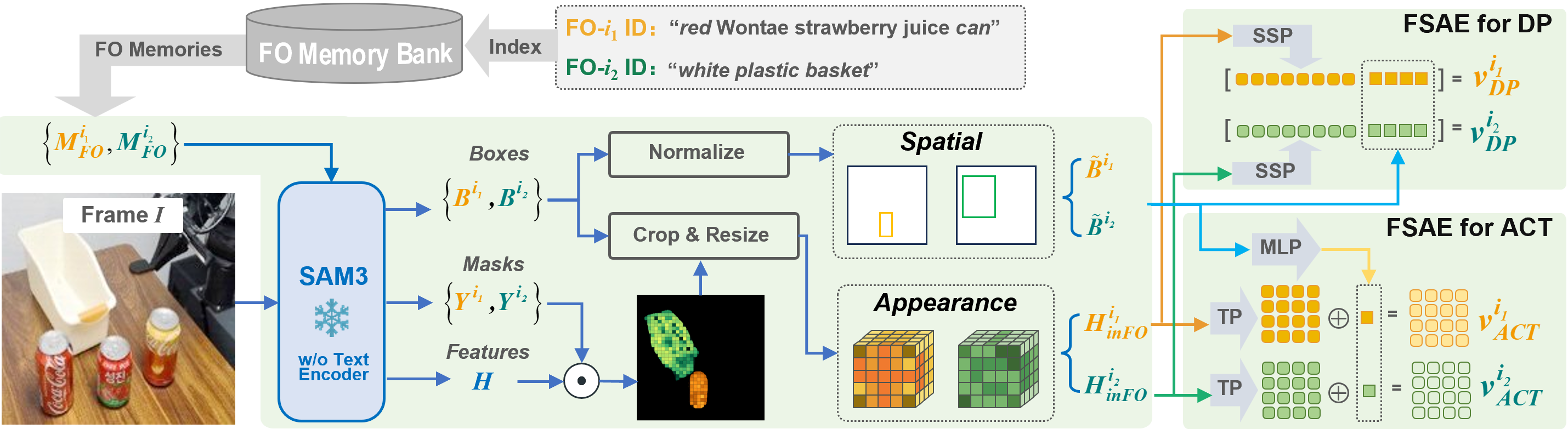}
\vspace{-20pt}
\caption{FOM-SAM3 guided policy with focused spatial-appearance encoding. Two FSAE versions are designed for DP and ACT, respectively. Here, SSP and TP stand for spatial softmax pooling and token projection, respectively.}
\label{fig:fsae}
\end{figure*}

\subsection{Focused Spatial-Appearance Encoding}
\label{sec:fsae}

The condition of a visuomotor policy should be focused on target FOs and
not distracted by irrelevant objects or background changes.
We propose a compact representation that concentrates on the spatial and
appearance states of target FOs, as illustrated in Fig.~\ref{fig:fsae}.

The explicit spatial state is represented by the normalized bounding box
$B^i
=
[c_x^i,c_y^i,w^i,h^i]$, which reflects the observed center, width, and height of the FO. The appearance state is represented by the in-FO features
$H^i_{inFO}$ from SAM3's feature pyramid network (FPN),
which retain richer visual cues of structure, pose, and manipulation progress.
The in-FO features are obtained by
\begin{equation}
H^i_{inFO}
=
\operatorname{Resize}
\left(
\operatorname{Crop}
\left(
H\odot Y^i,
B^i
\right)
\right),
\label{eq:fsae_focus}
\end{equation}
where $\odot$ is the Hadamard product.
Namely, we obtain the highest-level feature map $H$ of the SAM3's FPN, mask the non-FO pixels, crop the FO box, and resize it to a unified resolution of $16\times16$. Thus,
$H^i_{inFO}\in\mathbb{R}^{16\times16\times256}$ encodes the appearance cues
as an aligned feature map for arbitrarily-shaped FOs.

\subsection{Visual Conditioning for DP and ACT}
\label{sec:visual_condition}

The proposed FSAE is compatible with the two representative visuomotor
policy models DP~\cite{diffusion_policy} and ACT~\cite{act}.
The DP with a CNN structure utilizes a feature vector as the condition.
The ACT with a Transformer structure utilizes a sequence of tokens as the
condition.
Accordingly, FSAE further organizes the same spatial-appearance information into two distinct data structures for DP and ACT, respectively.

For DP, we project $H^i_{inFO}$ from 256 channels to 16 channels as
$\widetilde{H}^i_{inFO}$. Then the spatial softmax pooling (SSP) is applied to obtain
a spatial keypoint from each channel, as the original DP does. The resulting 16 keypoints
and the bounding box form the DP's condition vector:
\begin{equation}
v^i_{\mathrm{DP}}
=
\operatorname{Concat}
\big(
\operatorname{SSP}
(
\widetilde{H}^i_{inFO}
),
B^i
\big).
\label{eq:dp_condition}
\end{equation}

For ACT, we pool $H^i_{inFO}$ to the $8\times8$ resolution, then project it
to 512-D as the appearance embeddings
$\{\bar{H}^{i,k}_{inFO}\}_{k=1:64}$, to align with the token dimension of
ACT~\cite{act}.
Alongside, we project $B^i$ from 4-D to 512-D as the
spatial embedding $\bar{B}^i$.
The ACT visual tokens are formed by the sum of appearance embeddings and the
spatial embedding, as given by
\begin{equation}
v^i_{\mathrm{ACT}}
=
\left\{
\bar{H}^{i,k}_{inFO}
+
\bar{B}^i
\right\}_{k=1:64}.
\label{eq:act_condition}
\end{equation}

The execution pipeline of the FOM-SAM3 guided visuomotor policy is summarized
below.
First, a structured command is sent to the robot, for example,
``Action: DP\_Collect\_Can, Active FO: red Wontae strawberry juice can,
Passive FO: white plastic basket''.
The IDs of the active and passive FOs are used to retrieve their FO memory
tokens from the FO memory bank.
Second, the FO memory token sequences for the two target FOs are assigned to prompt SAM3.
If the two FOs are detected in the dual-view images, the segmentation results
are used to initialize the video object tracking of SAM3.
If the tracking confidence is too low, the FO memory-driven segmentation is rerun to relocalize
the FOs.
Third, the FO masks, boxes, and SAM3 features are fed to FSAE to obtain the policy conditions. The dual-view conditions of two FOs along with the robot
state sequence are used to generate a new action chunk.

\section{Experiments}
\label{sec:experiments}

\subsection{FOM-SAM3 Learning and Evaluation}
\label{sec:fom_learn}

\textit{1) FO-30 Dataset:} We constructed the FO-30 dataset from 30 physical FOs spanning four coarse categories, including can, cup, lid, and box. For each FO, we used a cellphone camera to casually capture up to 60 training images of it from various viewpoints on a plain background, then automatically obtained the mask annotations with the foreground cutout technique. The test dataset was built with the robot wrist camera and randomly placed FOs in different backgrounds, which comprises FO-30-S, with 600 standard test images, and FO-30-C,
with 400 challenging test images. The former has \(\sim\)2 FOs per image on average. The latter has \(\sim\)3 FOs per image on average and exhibits more occlusion cases. All the images have a 504$\times$504 resolution. Therefore, the test and training data differ in cameras, backgrounds, and clutter degrees, mainly to study FOM-SAM3's performance under various conditions within the dedicated operational domain.

\textit{2) Implementation Details:} For each FO, we optimized the objective in Eq.~\eqref{eq:fo_memory_loss} with AdamW using a learning rate of $1\times10^{-4}$,
weight decay of 0.1, gradient clipping at 1.0, batch size of 8, and a maximum of 30 epochs. Each training epoch contained 320 randomly composed images: 256 target-present scenes and 64 target-absent scenes. The learning time for a FO memory was $\sim$15 minutes on our NVIDIA RTX 5090 GPU.

\textit{3) Evaluation Metrics:}
Each test image was evaluated against all 30 FO queries, resulting in 30,000 image--FO pairs. FO-30-S comprises 1,268 target-present and 16,732 target-absent pairs, while FO-30-C comprises 1,192 target-present and 10,808 target-absent pairs. For target-present pairs, we report mIoU, Recall@0.5, and Prec@0.5, with the latter two metrics computed at a mask IoU threshold of 0.5. For target-absent pairs, we report the rejection rate, defined as the percentage of correctly rejected queries.

\begin{figure*}[t]
\centering
\begin{minipage}{\textwidth}
\includegraphics[width=\linewidth]{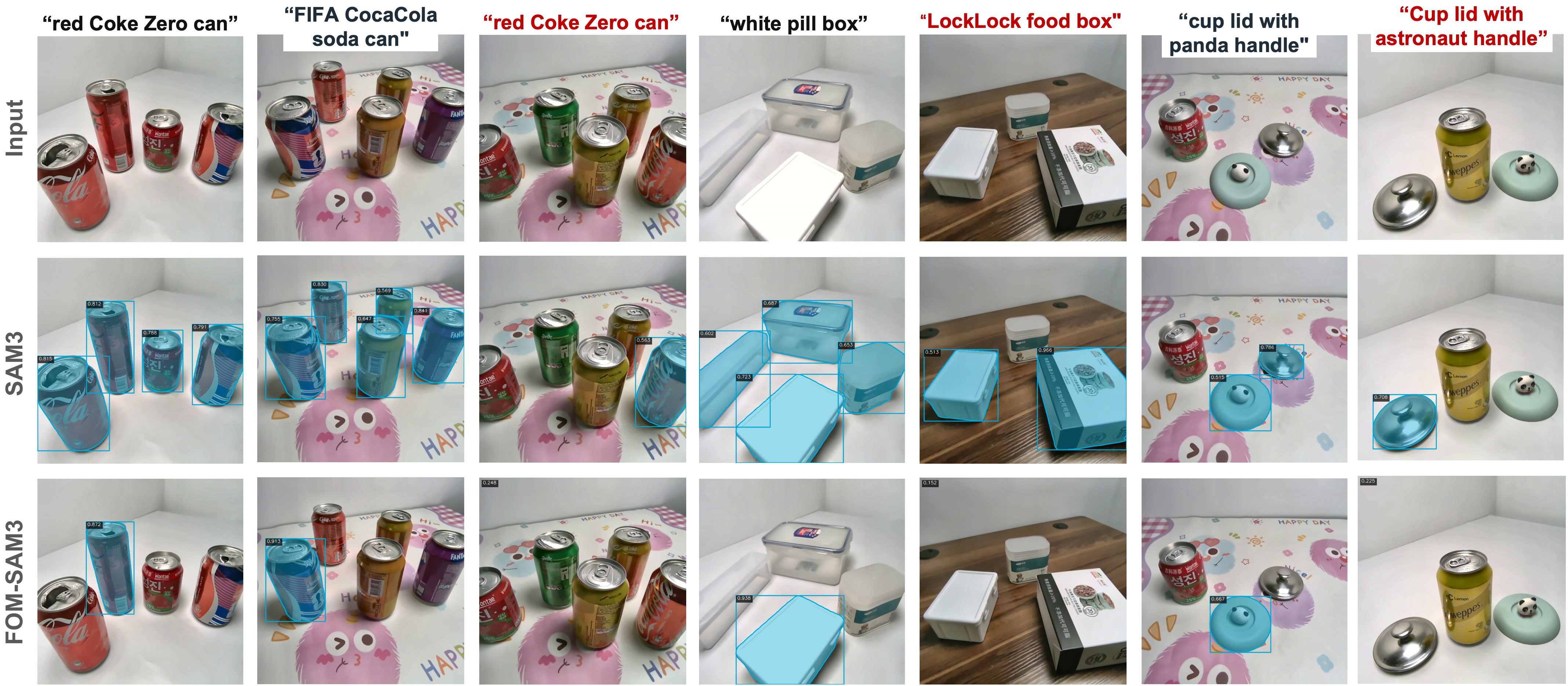}
\vspace{-20pt}
\caption{FO segmentation results on the FO-30 test set with the original SAM3 text prompting and the proposed FOM-SAM3. The red text indicates that the target FO is absent in the image.}
\label{fig:FO-seg}
\end{minipage}
\end{figure*}

\textit{4) Comparison Results on FO Perception Performance:}
We compared the FO perception performance of FOM-SAM3 with that of SAM3's original concept prompting. The latter used the same text prompt as the initial concept prompt in FO memory learning, which defined the color, category, brand, type, and material. We also compared FOM-SAM3 with three other prompt-based SAM3 post-training methods: Prompt-Tuning SAM3~\cite{ptsam}, Mask-to-Concept~\cite{m2c}, and Direct Token Optimization, while sharing the same loss function. Prompt-Tuning SAM3 introduced new learnable prompt tokens into SAM3. Mask-to-Concept learned an additive refinement vector for the initial prompt tokens to search within the original concept space. Direct Token Optimization directly updated the initial prompt tokens through gradient backpropagation and served as a baseline  for evaluating our strategy of keeping the initial prompt tokens frozen and learning a linear transformation.

As shown in Table~\ref{tab:fo_perception}, our FOM-SAM3 achieved the best FO perception performance on both the Standard and Challenge test sets, demonstrating superior FO segmentation accuracy, as well as a stronger ability to reject similar FOs within the same coarse-grained category. Compared with SAM3's original concept-level promptable segmentation, FOM-SAM3 with FO-wise memory learning improved mIoU, Recall@0.5, Prec@0.5, and Rejection by 22.86\%, 23.02\%, 17.17\%, and 12.06\% on the Standard test set, and by 25.79\%, 26.26\%, 33.34\%, and 6.96\% on the Challenge test set, respectively. Fig.~\ref{fig:FO-seg} presents qualitative comparisons of FO segmentation results between FOM-SAM3 and original SAM3. Furthermore, the proposed FO memory learning method outperformed the other three prompt-based post-training methods. We also evaluated the inference time of FOM-SAM3, which was $\sim$48.2ms, about 5ms shorter than that of original SAM3, because FOM-SAM3 bypassed the text prompt encoder during inference and used the learned memory tokens directly.

\begin{table}[t]
\centering
\caption{\textbf{FO Perception Performances Based on Frozen SAM3 (\%)}}
\label{tab:fo_perception}
\vspace{-6pt}
\scriptsize
\setlength{\tabcolsep}{3.8pt}
\begin{tabular}{lcccc}
\toprule
\textbf{Method}
& \textbf{mIoU$\uparrow$}
& \textbf{Recall@0.5$\uparrow$}
& \textbf{Prec@0.5$\uparrow$}
& \textbf{Rejection$\uparrow$} \\
\midrule
\multicolumn{5}{c}{\textit{FO-30-S: Standard}}\\
\midrule
Original Concept Prompt~\cite{sam3}
& 70.18 & 71.85 & 76.30 & 81.04 \\
Prompt-Tuning SAM3~\cite{ptsam}
& 68.22 & 69.79 & 78.32 & 84.25 \\
Mask-to-Concept~\cite{m2c}
& 84.85 & 86.99 & 78.12 & 82.74 \\
Direct Token Optimization
& 84.77 & 86.83 & 78.03 & 82.81 \\
\rowcolor{bestblue}
\bestmethod{FOM-SAM3 (Ours)}
& \bestvalue{93.04}
& \bestvalue{94.87}
& \bestvalue{93.47}
& \bestvalue{93.10} \\
\midrule
\multicolumn{5}{c}{\textit{FO-30-C: Challenge}}\\
\midrule
Original Concept Prompt~\cite{sam3}
& 57.46 & 59.06 & 53.66 & 85.58 \\
Prompt-Tuning SAM3~\cite{ptsam}
& 53.19 & 54.70 & 58.06 & 87.37 \\
Mask-to-Concept~\cite{m2c}
& 69.63 & 71.73 & 57.27 & 86.90 \\
Direct Token Optimization
& 68.84 & 70.89 & 57.40 & 86.94 \\
\rowcolor{bestblue}
\bestmethod{FOM-SAM3 (Ours)}
& \bestvalue{83.25}
& \bestvalue{85.32}
& \bestvalue{87.00}
& \bestvalue{92.54} \\
\bottomrule
\end{tabular}
\end{table}

\subsection{FO Manipulation Experiments with Real Robot}
\label{sec:robot}
\begin{figure*}[t]
\centering
\includegraphics[width=\linewidth]{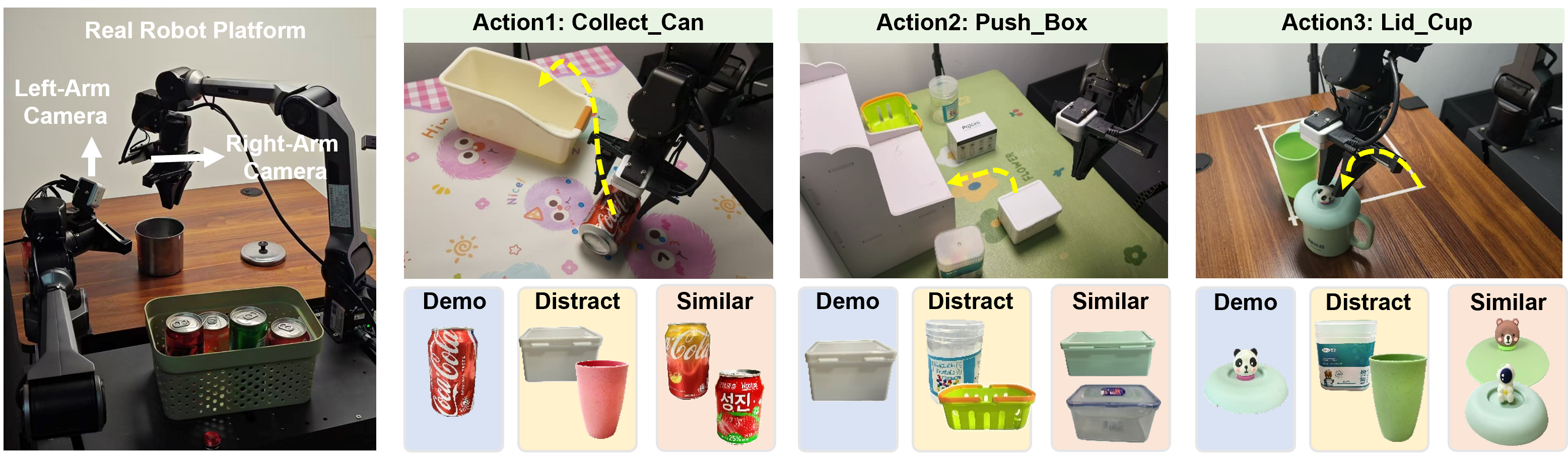}
\vspace{-20pt}
\caption{Real-robot platform, three actions, and experimental objects. DemoFO is the object used in teleoperated demonstration. Distractors are dissimilar objects from different coarse categories. SimilarFOs are visually confusing objects of the same category.}
\label{fig:robot_tasks}
\end{figure*}

\textit{1) Platform and Tasks:}
As shown in Fig.~\ref{fig:robot_tasks}, the experiment platform had two AgileX PiPER-X robot arms. Each robot arm was equipped with a two-finger gripper and an Intel RealSense D405 wrist camera. The right arm provided a first-person view and performed the manipulation, while the left arm maintained a fixed pose throughout the operation solely to provide a third-person view. For better real-time performance, raw camera frames were center-cropped to $960\times720$ pixels and resized to $336\times336$ pixels. We evaluated the action policies on three tasks: 1) pick the target can and place it into a storage box from above (Collect\_Can); 2) push the target box into the opening at the bottom of a shelf (Push\_Box); 3) put the target lid on a cup (Lid\_Cup). An AgileX PIKA SENSE teleoperation kit was utilized to collect human demonstration data.

\textit{2) Implementation Details:} The data collection frequency in demonstration and the policy inference frequency in execution were both 10\,Hz. loop frequency. The motion command publishing frequency was 50\,Hz after online trajectory interpolation. For each task, 120 human demonstrations were collected for visuomotor policy learning. For our FOM-SAM3-guided DP, the two-FO visual states from two views, the two-step robot state history, and the 128-D diffusion-time embedding formed a 292-D FiLM condition to guide the 1-D UNet policy. The prediction horizon and execution horizon were 24 and 16, respectively. The policy was trained for 150k steps using Adam with a batch size of 16, learning rate of $1\times10^{-3}$, and weight decay of
$1\times10^{-5}$. We used 100 diffusion timesteps during training and 35 denoising steps
during deployment. For our FOM-SAM3-guided ACT, the visual tokens of two FOs from two views, the latent token, and the robot state token form the input of the Transformer encoder. The action chunk size was 30. The execution step after temporal ensemble was 1. The policy
was trained for 100k steps using AdamW with a learning rate of $1\times10^{-4}$,
weight decay of $1\times10^{-3}$, and batch size of 16.

\begin{figure}[t]
\centering
\begin{minipage}{0.49\textwidth}
\includegraphics[width=\linewidth,trim={5mm 15mm 0mm 0mm},
  clip]{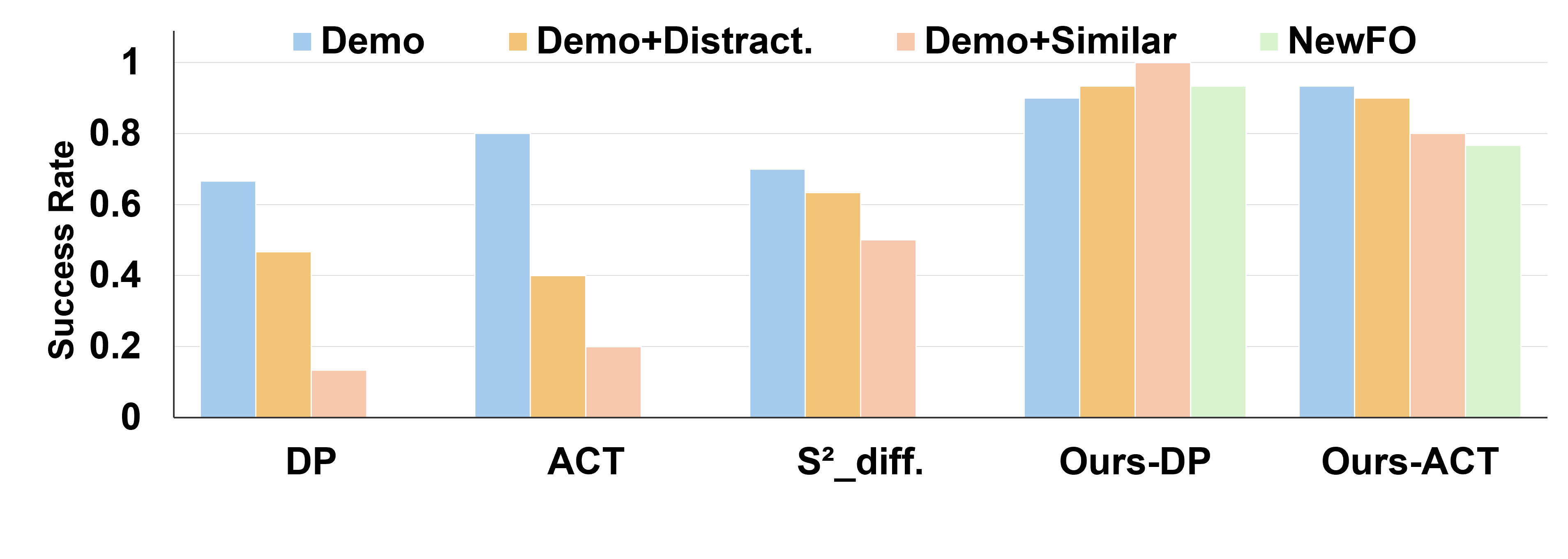}
\setlength{\abovecaptionskip}{-12pt}
\caption{Success rates of different methods in all real-robot evaluations.}
\label{fig:success-rates}
\end{minipage}
\end{figure}

\begin{table}[!t]
\centering
\caption{\textbf{Success Rates of DemoFO Manipulation}}
\label{tab:robot_main}
\vspace{-6pt}
\scriptsize
\setlength{\tabcolsep}{3pt}
\begin{tabularx}{\columnwidth}{@{}l*{4}{Y}@{}}
\toprule
\textbf{Method}
& \makecell{\textbf{Collect\_Can}}
& \makecell{\textbf{Push\_Box}}
& \makecell{\textbf{Lid\_Cup}}
& \textbf{Overall} \\
\midrule
DP~\cite{diffusion_policy}
& 15/30 & 13/30 & 10/30 & 38/90 \\
S$^2$-Diff.~\cite{s2diffusion}
& 22/30 & 19/30 & 14/30 & 55/90 \\
\rowcolor{bestblue}
\bestmethod{Ours-DP}
& \bestvalue{30/30}
& \bestvalue{27/30}
& \bestvalue{28/30}
& \bestvalue{85/90} \\
\midrule
ACT~\cite{act}
& 14/30 & 18/30 & 10/30 & 42/90 \\
\rowcolor{bestblue}
\bestmethod{Ours-ACT}
& \bestvalue{26/30}
& \bestvalue{28/30}
& \bestvalue{25/30}
& \bestvalue{79/90} \\
\bottomrule
\end{tabularx}
\end{table}

\begin{table}[!t]
\centering
\caption{\textbf{Success Rates of NewFO Manipulation with Shared Policy}}
\label{tab:within_type_reuse}
\vspace{-6pt}
\scriptsize
\setlength{\tabcolsep}{5pt}
\begin{tabularx}{0.98\columnwidth}{@{}l*{4}{Y}@{}}
\toprule
\textbf{Method} & \textbf{Collect\_Can} & \textbf{Push\_Box} & \textbf{Lid\_Cup} & \textbf{Overall} \\
\midrule
\rowcolor{bestblue}\bestmethod{Ours-DP} & \bestvalue{10/10} & \bestvalue{10/10} & \bestvalue{8/10} & \bestvalue{28/30} \\
\rowcolor{bestblue}\bestmethod{Ours-ACT} & \bestvalue{7/10} & \bestvalue{8/10} & \bestvalue{8/10} & \bestvalue{23/30} \\
\bottomrule
\end{tabularx}
\end{table}

\textit{3) Evaluation Setup:}
We designed three evaluation settings with increasing levels of difficulty. Under each evaluation setting, an action was \emph{uninterruptedly} performed for 10 rollouts. A rollout was counted as
successful only if the robot selected the target FO and completed
the full task. A completed action with wrong FO selection was also counted as a failure. Each method was evaluated on all three tasks under all three settings. Predefined positions were marked on the desk to ensure test consistency across methods. Thus, each method was evaluated for 90 rollouts in total. The three evaluation settings are as follows.

\emph{DemoFO:}
To evaluate the action performance across different object positions and orientations, 
the FO used in the human demonstrations was designated as the manipulation target (DemoFO). No distractors or confusing FOs are involved in the scene. 

\emph{DemoFO$+$Distractors:}
The manipulation target was DemoFO. To evaluate robustness to scene distractors, two objects from other coarse-grained categories were placed beside the target FO in the scene.

\emph{DemoFO$+$SimilarFOs:}
The manipulation target was DemoFO. To evaluate the ability to distinguish similar FOs within the same coarse-grained category, two similar FOs were placed beside the target FO in the scene.

\emph{NewFO:}
To further test whether our FOM-SAM3-guided policy could be applied to new FOs within the same coarse-grained category, while sharing the policy trained with DemoFO. We switched the memory prompt tokens of FOM-SAM3 to those of another new FO (NewFO) that was unseen during policy training. The DemoFO was in turn placed beside the NewFO as a confusing object. We conducted five rollouts for each of two NewFOs.

\begin{figure*}[t]
    \centering
    \includegraphics[width=\textwidth]{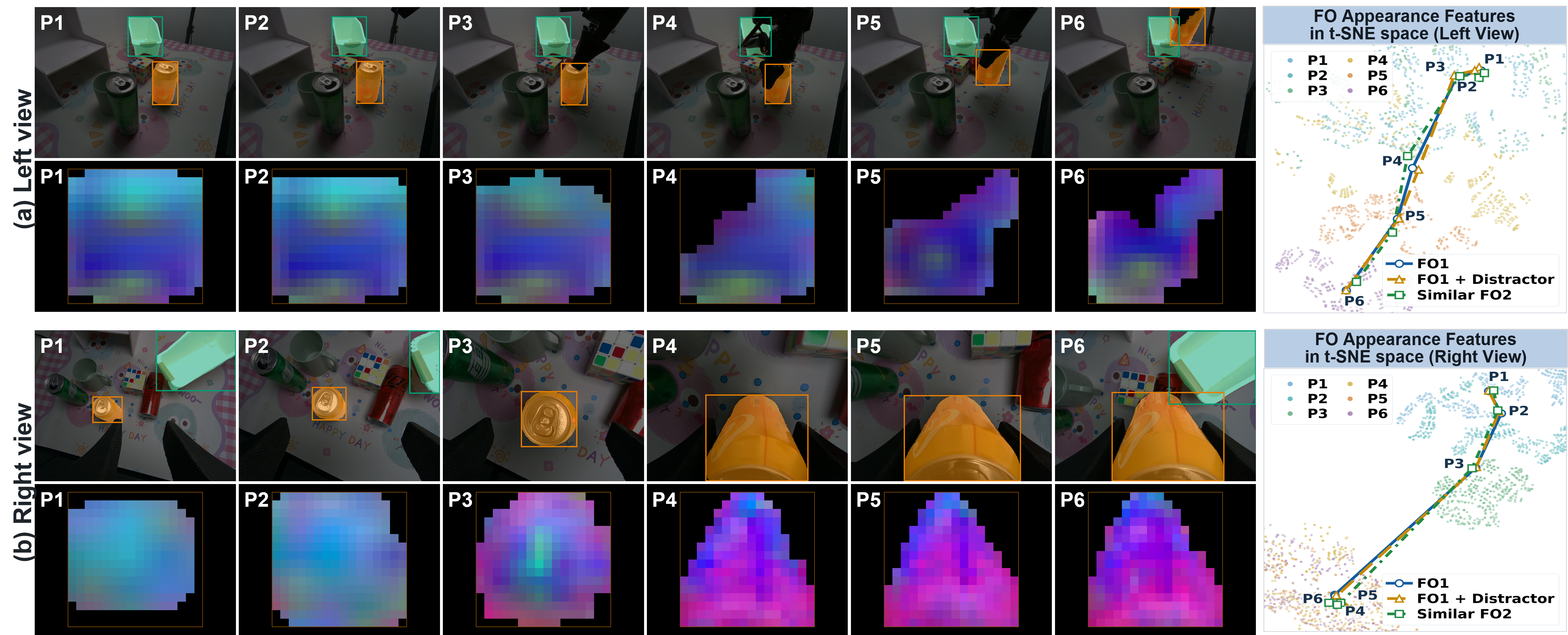}
    \caption{Visualization of FSAE representations in the left view (a) and the right view (b). In each subfigure, the first row shows the boxes and mask given by FOM-SAM3. The second row visualizes the in-FO appearance features after unified PCA feature dimensionality reduction. The waypoint number on the top-left corner of each picture corresponds to that in the t-SNE map on the far right.}
    \label{fig:vis-fsae}
\end{figure*}

\textit{4) Comparison of DemoFO Manipulation Performances:}
We used the vallina DP ~\cite{diffusion_policy}
and ACT~\cite{act} under the dual-view configuration as baselines. We also
compared with S$^2$-Diffusion~\cite{s2diffusion}, 
expanding its original implementation from single-view to dual-view. Our FOM-SAM3 guided DP (Ours-DP) shared the same policy architecture and training configuration with DP and S$^2$-Diffusion, while the only difference lay in the visual conditioning approach. Similarly, our FOM-SAM3 guided ACT (Ours-ACT) shared the same policy architecture and training configuration with ACT. Table~\ref{tab:robot_main} reports the success rates of the different methods. Our FOM-SAM3 guided DP achieved the best overall success rate, followed by our FOM-SAM3 guided ACT. 

\emph{Effect of Focused Visual Conditioning:} As shown in Fig.~\ref{fig:success-rates}, when only the manipulation target DemoFO was in the scene, all the methods achieved success rates of 0.66$\sim$0.93, among which our methods performed the best. After introducing visual distractors, the success rates of DP and ACT decreased significantly because of the limitation of the scene-level visual conditioning. In contrast, S$^2$-Diffusion and our methods demonstrated robustness to distractors in the scene. 

\emph{Effectiveness of FOM-SAM3:} When confusing SimilarFOs were involved in the scene, the success rates of DP, ACT, and S\(^2\)-Diffusion all fell to 0.13$\sim$0.5, because both the scene- and category-level visual perception could not distinguish the target FO from other similar FOs. In comparison, our methods maintained the high success rates due to the fine-grained guidance of our FOM-SAM3.

\textit{5) NewFO Manipulation Performance with Shared Policy:}
As shown in Table~\ref{tab:within_type_reuse}, our FOM-SAM3-guided DP and ACT achieved success rates of 93.3\% and 76.7\%, respectively. These results demonstrated the reusability of our method across novel FOs within the same coarse-grained category. The higher success rate of Ours-DP over Ours-ACT was due to DP’s ability to generate multimodal actions. Although ACT modeled multimodal action distributions during training, its inference was still deterministic and might suffer from alignment failures.

\textit{6) Visualization of Explicit Spatial and Implicit Appearance
Representation in FSAE:}
We collected the dual-view images at six predefined robot waypoints (P1$\sim$P6) for three rounds. In rounds 1 and 3, a ``Coca-Cola lemon soda can'' (FO1) and ``Schweppes grape soda can'' (Similar FO2) were used as the manipulation target, while no distractors were involved. In round 2, FO1 was used as the manipulation target, and several distractors were involved (FO1$+$Distractor). Fig.~\ref{fig:vis-fsae}(a) and (b) illustrate the FO perception results in round 2 from the left and right views, respectively. First, a bounding box explicitly indicates the center position, width, and height of the FO. Moreover, its scale and aspect ratio also reflect the variation in depth and orientation. Second, the segmentation masks were further used to remove the non-FO pixels within and outside the boxes. Third, from the visualized in-FO appearance features after PCA dimensionality reduction, we found that the in-FO features were highly correlated with the structure and pose of FO, providing implicit higher-dimensional visual cues of the manipulation state and progress.

As shown in the rightmost plots of Fig.~\ref{fig:vis-fsae}(a) and (b), we compared the in-FO appearance feature distributions in rounds 1$\sim$3 by projecting them into t-SNE spaces for the left and right views, respectively~\cite{tsne}. The pale-colored points display the features of all the in-FO pixels. The centroids of the features in each round are marked and connected as a trajectory. As we can see, when the robot approached the target FO from P1$\sim$P3, the image appearances presented limited changes, so the pixel features and centroids shifted in a relatively small range. When the robot gripped and lifted the FO from P4$\sim$P6, the appearance features shifted significantly in the t-SNE space. Notably, in the left-view images at P1$\sim$P3 and the right-view images at P4$\sim$P6, the FO appearances remained approximately unchanged, so their in-FO features were also clustered together in the t-SNE space. Therefore, the appearance representation of FSAE could effectively indicate the variation and similarity of FO state in an entire manipulation process. Meanwhile, by comparing the three trajectories, we found that visual distractors and replacing the target with a SimilarFO only
slightly perturbed the FSAE result.

\subsection{Limitations}
\label{sec:limitation}
We mainly observed three limitations in this work. First, the FO perception might fail when the discriminative features are invisible. For example, we cannot identify the brand and flavor of a can from its top view. However, view angle changes and severe occlusion are inevitable in real applications. Second, the FSAE method excludes irrelevant objects, which might limit the potential obstacle avoidance ability of a visuomotor policy. Third, incorporating the foundation model SAM3 as a visual encoder provides object-focused visual information but increases the latency from frame input to motion command.

\section{Conclusion}
We presented a FOM-SAM3-guided visuomotor policy framework that combines persistent FO memory with focused visual conditioning for FO manipulation. FOM-SAM3 learns and stores FO-specific memory tokens, enabling a registered FO to be localized and distinguished from visually similar objects. FSAE then combines the target FO's appearance features and explicit spatial states to guide the action policy. Experiments on the FO-30 dataset and three real-robot tasks demonstrate improved FO perception and manipulation success in the presence of distractors and confusing FOs. The framework supports policy reuse across new FOs within the same coarse category by changing only the retrieved FO Memory. These results highlight the value of persistent object memory for extending learned manipulation skills to additional familiar objects.

\begingroup
\raggedbottom
\balance
\sloppy
\hbadness=10000
\setlength{\emergencystretch}{3em}
\Urlmuskip=0mu plus 1mu\relax
\bibliographystyle{IEEEtran}
\bibliography{references}
\endgroup

\end{document}